\documentclass[lettersize,journal]{IEEEtran}
\usepackage{amsmath,amsfonts}
\usepackage{algorithmic}
\usepackage{array}
\usepackage{textcomp}
\usepackage{stfloats}
\usepackage{url}
\usepackage{verbatim}
\usepackage{graphicx}
\def\BibTeX{{\rm B\kern-.05em{\sc i\kern-.025em b}\kern-.08em
    T\kern-.1667em\lower.7ex\hbox{E}\kern-.125emX}}
\usepackage{balance}

\usepackage{color}
\usepackage[utf8]{inputenc}
\usepackage{multirow}
\DeclareUnicodeCharacter{2212}{-}
\usepackage{booktabs}

\usepackage{times}
\usepackage{epsfig}
\usepackage{graphicx}
\usepackage{subfigure}

\newcommand\blfootnote[1]{%
  \begingroup
  \renewcommand\thefootnote{}\footnote{#1}%
  \addtocounter{footnote}{-1}%
  \endgroup
}

\begin{document}
\title{Learning Branched Fusion and Orthogonal Projection for Face-Voice Association}
\author{Muhammad Saad Saeed$^{1\ast}$, 
Shah Nawaz$^{2\ast \dagger}$, 
Muhammad Haris Khan$^{3}$, 
Sajid Javed$^{4}$,\\
Muhammad Haroon Yousaf$^{1}$, 
Alessio {Del Bue}$^{2,5}$ \\ 
$^{1}$Swarm Robotics Lab (SRL)-NCRA, University of Engineering and Technology Taxila,
$^{2}$Pattern Analysis \& Computer Vision (PAVIS) - Istituto Italiano di Tecnologia (IIT),
$^{3}$Mohamed Bin Zayed University of Artificial Intelligence,
$^{4}$Khalifa University of Science and Technology,
$^{5}$Visual Geometry \& Modelling (VGM) - Istituto Italiano di Tecnologia (IIT)
}

\markboth{Journal of \LaTeX\ Class Files,~Vol.~18, No.~9, September~2020}%
{How to Use the IEEEtran \LaTeX \ Templates}

\maketitle

\begin{abstract}
\blfootnote{$\ast$ Equal contribution}
\blfootnote{$\dagger$ Current Affiliation: Deutsches Elektronen-Synchrotron}
Recent years have seen an increased interest in establishing association between faces and voices of celebrities leveraging audio-visual information from YouTube. 
Prior works adopt metric learning methods to learn an embedding space that is amenable for associated matching and verification tasks. 
Albeit showing some progress, such formulations are, however, restrictive due to dependency on distance-dependent margin parameter, poor run-time training complexity, and reliance on carefully crafted negative mining procedures.
In this work, we hypothesize that an enriched representation coupled with an effective yet efficient supervision is important towards realizing a discriminative joint embedding space for face-voice association tasks.
To this end, we propose a light-weight, plug-and-play mechanism that exploits the complementary cues in both modalities to form enriched fused embeddings and clusters them based on their identity labels via orthogonality constraints.
We coin our proposed mechanism as fusion and orthogonal projection (FOP) and instantiate in a two-stream network. The overall resulting framework is evaluated on VoxCeleb1 and MAV-Celeb datasets with a multitude of tasks, including cross-modal verification and matching. 
Results reveal that our method performs favourably against the current state-of-the-art methods and our proposed formulation of supervision is more effective and efficient than the ones employed by the contemporary methods.
In addition, we leverage cross-modal verification and matching tasks to analyze the impact of multiple languages on face-voice association. Code is available: \url{https://github.com/msaadsaeed/FOP}
%
\end{abstract}


\section{Introduction}
\begin{figure*}
\centering
\includegraphics[scale=0.22]{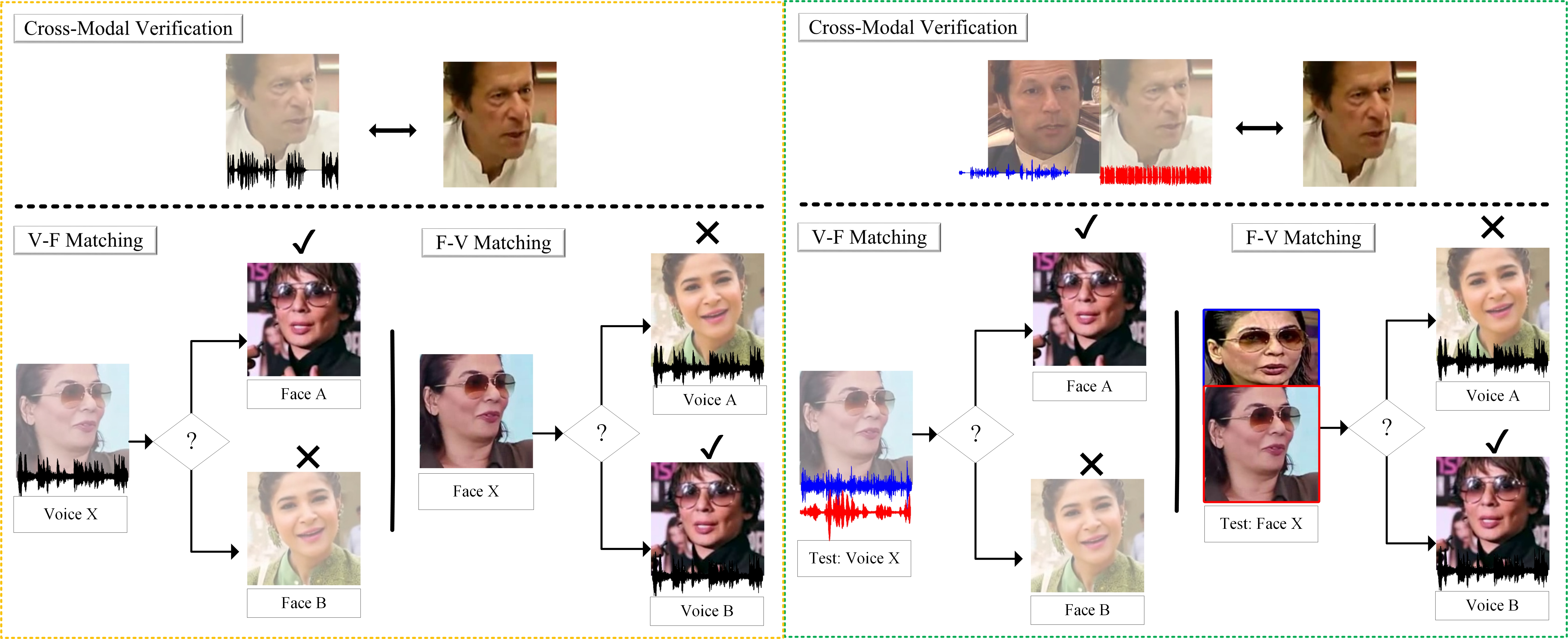}
   \caption{Face-voice association is established in cross-modal verification and matching tasks. The goal of cross-modal verification task is to verify if a clip of a voice and face image belong to the same identity. In a cross-modal matching task, Voice-Face(V-F): given a clip of a voice and two or more face images, select the face image that belongs to the voice. (Face-Voice) F-V: given an image of a face, determine the respective voice.
   In addition, we analyze the impact of multiple of languages on cross-modal verification and matching tasks.} 
\label{fig:mfigure}
\end{figure*}

\IEEEPARstart{I}{t} is a well-studied and understood fact that humans can associate voices and faces of people because the neuro-cognitive pathways for voices and faces share same structure~\cite{kamachi2003putting,belin2004thinking}. Recently, Nagrani et al.~\cite{nagrani2018seeing,nagrani2018learnable,nagrani2017voxceleb} introduced the face-voice association task to vision community with the creation of a large-scale audio-visual dataset, comprising faces and voices of $1,251$ celebrities. Since then, the face-voice association task has gained increased research interest~\cite{horiguchi2018face,kim2018learning,nagrani2018learnable,nagrani2018seeing,nawaz2019deep,wen2021seeking,wen2018disjoint,zhu2022unsupervised}. In addition, we observe the creation of new audio-visual datasets for studying novel tasks~\cite{kim2018learning,nawaz2021cross,wu2022cross,zhu2022celebvhq}. For example, fundamental face-voice association tasks are studied to analyze the impact of multiple languages, see Fig.~\ref{fig:mfigure}.
%

Most existing works ~\cite{kim2018learning,nagrani2018seeing,nagrani2018learnable,nawaz2021cross} tackle face-voice association as a cross-modal biometric task. The two prominent challenges in developing an effective method for this task are learning of a common yet discriminative embedding space, where instances from two modalities are sufficiently aligned and instances of semantically similar identities are nearby. Often separate networks for face and voice modalities are leveraged to obtain the respective feature embeddings and contrastive or triplet loss formulations are employed to construct this embedding space. 

Although showing some effectiveness in this task, such loss formulations, however, are restrictive in following ways. First, they require tuning of a margin hyperparameter, which is hard as the distances between instances can alter significantly while training. Secondly, the run-time training complexity for contrastive and triplet losses are $\mathcal{O} (n^2)$ and $\mathcal{O} (n^3)$, respectively, where $n$ is the number of available instances for a modality. Finally, to mitigate the high run-time training complexity challenge, different variants of carefully crafted negative mining strategies are used, which are both time-consuming and performance sensitive. 

A few methods e.g., \cite{nawaz2019deep} have attempted to replace the contrastive/triplet loss formulations by utilizing auxiliary identity centroids~\cite{wen2016discriminative}. The training process alternates between the following two steps: 1) clustering embeddings around their identity centroids and pushing embeddings away from all other identity centroids, and 2) updating these centroids using the mini-batch instances. Such centroid based losses are used with traditional classification loss (i.e. softmax cross-entropy (CE)). However, their co-existence is unintuitive and ineffective because the former promotes margins in Euclidean space whereas latter implicitly achieves separability in the angular domain.

In this work, we hypothesize that an enriched unified feature representation, encompassing complementary cues from both modalities, alongside an effective yet efficient supervision formulation is crucial towards realizing a discriminative joint embedding space for improved face-voice association. To this end, we propose a light-weight, plug-and-play mechanism that exploits the best in both modalities through fusion and semantically aligns fused embeddings with their identity labels via orthogonality constraints. We instantiate our proposed mechanism in the two-stream pipeline, which provides face and voice embeddings, and the resulting overall framework is an effective and efficient approach for face-voice matching and verification tasks.



We summarize our key contributions as follows. 1) We propose to harness the complementary features from both modalities in forming enriched feature embeddings, that are consistent with semantics of identity, thereby allowing improved identity recognition. 2) We propose to impose orthogonality constraints on the fused embeddings. They are not only coherent with the angular characteristic of the commonly employed classification loss but are very efficient as they operate directly on mini-batches. 3) Experimental results on large-scale VoxCeleb$1$ \cite{nagrani2017voxceleb} show the effectiveness of our method on both face-voice verification and matching tasks. Further, we note that our method performs favourably against the existing state-of-the-art methods. 4) We perform a thorough ablation study to analyze the impact of different components and empirically show that the proposed supervision formulation for face-voice retrieval is more effective and efficient than the ones employed by the contemporary works. 

A preliminary version of this work was published in ~\cite{saeed2022fusion}. In addition, the current manuscript makes following new contributions. 
We study the effectiveness of our approach in the challenging setting of establishing face-voice association across multiple languages spoken by the same set of persons.
The experiments are performed on Multilingual Audio-Visual (MAV-Celeb) dataset~\cite{nawaz2021cross}.
Our results indicate that, when training on language `A' and testing on \textit{unheard} language `B', the performance in face-voice association tasks deteriorates notably, and this performance is significantly poor when the same model (i.e. trained on language `A') is tested on \textit{heard} language `A'.
However, the performance in the former setting (\textit{unheard} language `B') is still better than random, which is of significance considering the challenging nature and the configuration of the proposed evaluation protocol.
Our analysis reveal that this performance drop is due to notable mismatch between the feature distributions of the two languages, typically known as the domain shift problem.
Finally, to our knowledge, we provide a first rigorous comparison of loss functions employed in existing face-voice association methods by utilizing the same underlying branch architecture, input features, and evaluation protocols on VoxCeleb$1$ dataset (see Fig.~\ref{fig:comparison}).

\section{Related Work}
We begin by discussing face-voice association from a cognitive neuroscience and psychological perspective, and then detail overview of related work on face-voice (F-V) association.

\noindent \textbf{Cognitive Psychological Perspective.}
Studies have shown that humans are capable of associating voices of unknown identities to their faces. 
In~\cite{mcallister1993eyewitnesses, schweinberger2007hearing}, authors performed a studies to show that humans memorize and recall voices of persons which they have seen (learned) previously.
In a similar study ~\cite{kamachi2003putting,belin2004thinking}, authors reveal the presence of equivalent amount of information about an identity from audio and visual streams. 
Hence, we can say that multimodality can improve the perception capability of humans~\cite{munhall1998moving}. In ~\cite{munhall1998moving}, authors show that human perceptual accuracy improves when visual modality supplements the auditory modality in a noisy environment. 
We note that these studies have been conducted previously under the cognitive psychological perspective in detail and have recently inspired the vision community~\cite{nagrani2018seeing,nagrani2018learnable}.

\noindent \textbf{Face-voice Association.}
The work of Nagrani et al.~\cite{nagrani2018seeing} leveraged audio and visual information to establish an association between faces and voices in a cross-modal biometric matching task.
Similarly, some recent metric learning work~\cite{kim2018learning,nagrani2018learnable,wen2021seeking,nawaz2021cross,tao20b_interspeech,chen2022self} introduced joint embeddings to establish correlation between face and voice of an identity. These methods extract audio and face embeddings and then minimize the distance between embeddings of same identities while maximize the distance among embeddings from different ones.  
Zheng et al.~\cite{zheng2021adversarial} proposes Adversarial-Metric Learning model combing metric and adversarial learning. The aim of adversarial learning is to generate modality independent representation while the metric learning learns a metric for audio-visual cross-modal matching and retrieval tasks.
Wen et al.~\cite{wen2018disjoint} presented a disjoint mapping network to learn a shared representation for audio and visual information by mapping them individually to common covariates (gender, nationality, identity) without needing to construct pairs or triplets at the input. 
Similarly, Nawaz et al.~\cite{nawaz2019deep} extracted audio and visual information with a single stream network to learn a shared deep latent representation, leveraging identity centroids to eliminate the need of pairs or triplets~\cite{nagrani2018learnable,nagrani2018seeing}. 
Both Wen et al.~\cite{wen2018disjoint} and Nawaz et al.~\cite{nawaz2019deep} show that effective face-voice representations can be learned without pairs or triplets formation.
More recently, Ning et al.~\cite{ning2021disentangled} proposed to disentangle the alignable latent identity factors and nonalignable the modality dependent factors for cross-modal biometric matching.

In contrast to existing, our method differs in following respects. First, it proposes to construct enriched embeddings via exploiting complementary cues from the embeddings of both modalities through a attention-based fusion. Second, it clusters the embeddings of same identity and separates embeddings of different identities via orthogonality constraints. The instantiation of both proposals in a two-stream pipeline results in an effective and efficient face-voice association framework.
 




\begin{figure*}[!htp]

\includegraphics[scale=0.54]{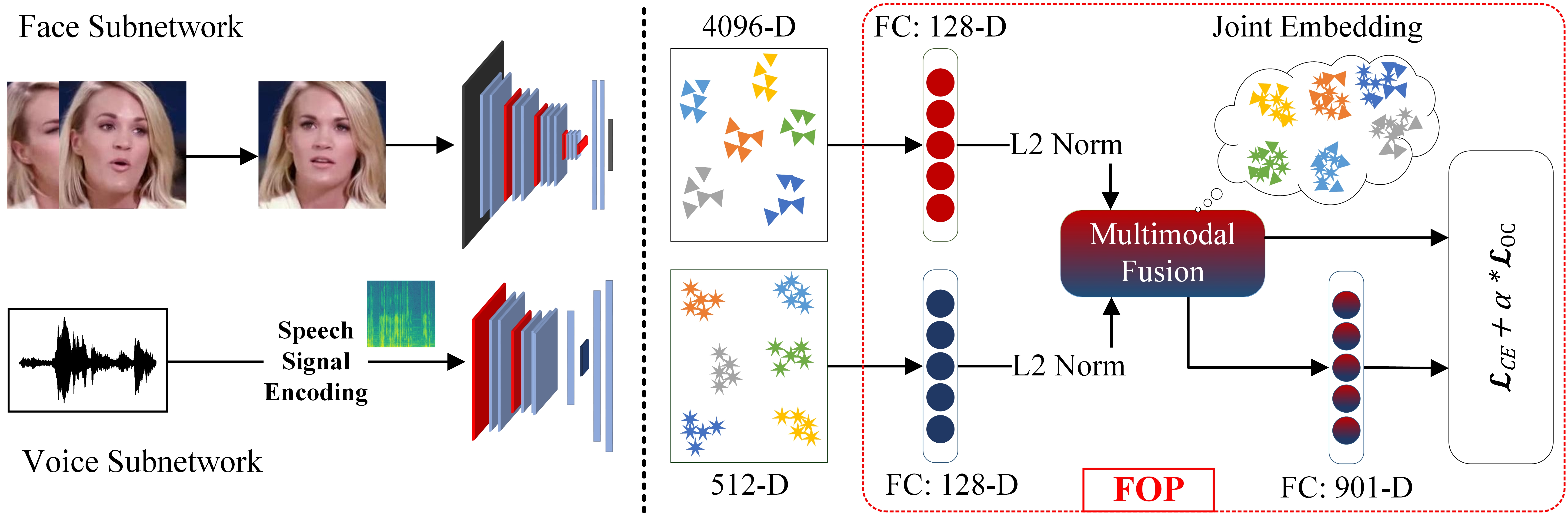}
\includegraphics[scale=0.7]{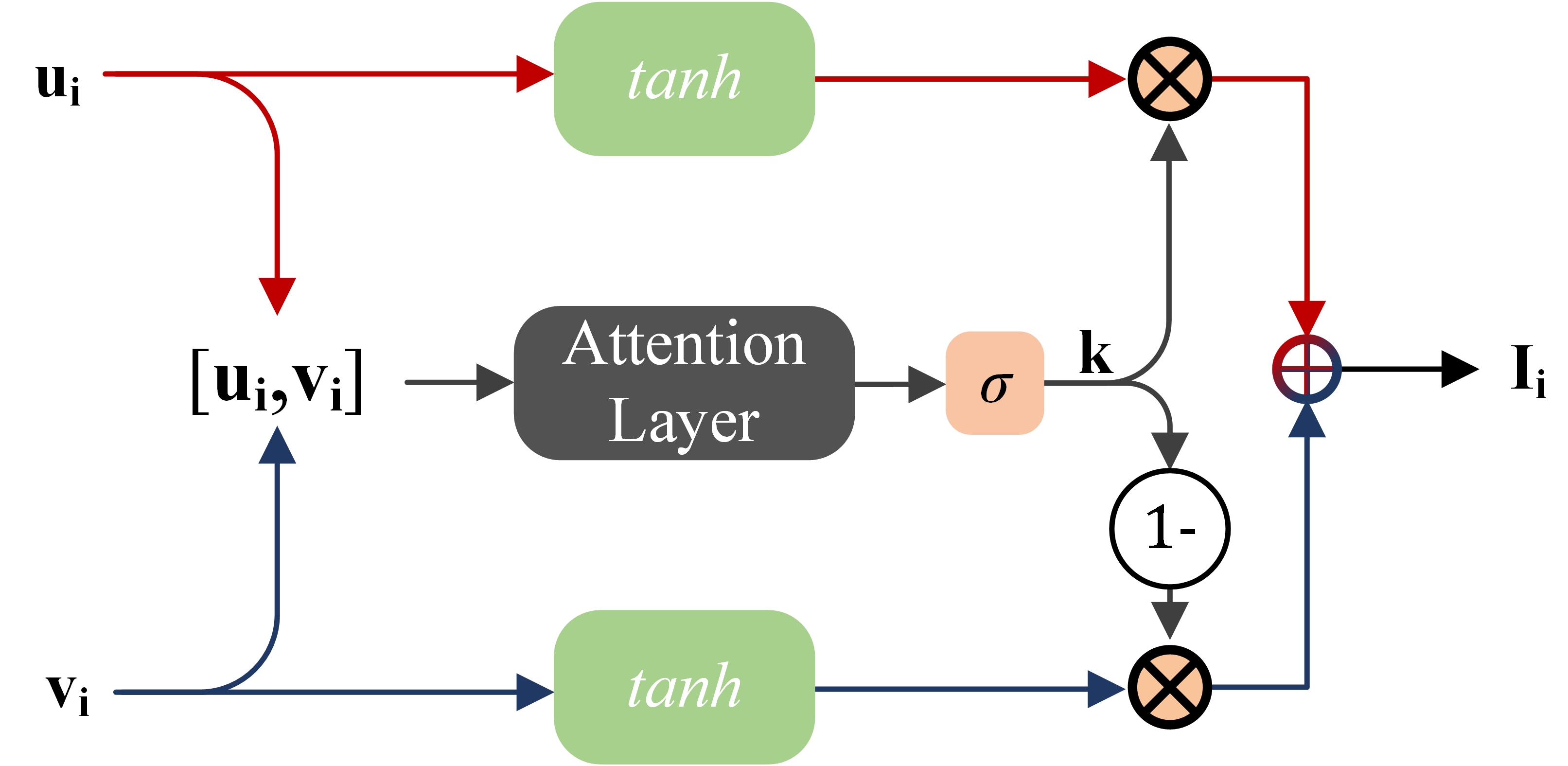}
   \caption{Architecture of our method. Fundamentally, it is a two-stream pipeline which generates face and voice embeddings. We believe that an enriched embedding coupled with an effective yet efficient supervision is key in learning a discriminative joint embedding space for improved face-voice association. To this end, we propose a light-weight, plug-and-play mechanism, dubbed as fusion and orthogonal projection (FOP) (shown in dotted red box). It is capable of exploiting complementary cues in both modalities and clustering the enriched fused embeddings based on their identity labels via imposing orthogonality constraints with the traditional classification loss (i.e. softmax cross-entropy).
   }
\label{fig:overall_framework_fusion}
\end{figure*}

\section{Overall Framework}
\label{section:overall_framework}
Our aim is to learn a discriminative joint face-voice embedding that is amenable to a multitude of tasks such as face-voice/voice-face matching, verification and retrieval. To actualize this, we develop a new framework for cross-modal face-voice association (See Fig.~\ref{fig:overall_framework_fusion}) that is fundamentally a two-stream pipeline (sec.~\ref{subsection:Preliminaries}) and features a light-weight module that exploits complementary cues from both face and voice embeddings and facilitates discriminative identity mapping via orthogonality constraints (sec.~\ref{subsection:Learning_Discriminative_Joint_Embeddings}).

\subsection{Preliminaries}
\label{subsection:Preliminaries}
\noindent \textbf{Problem Settings.}
Without the loss of generality, we consider cross-modal retrieval of bimodal data, i.e., for face and voice. Given that we have N instances of face-voice pairs, $\mathcal{D}=\{(x_{i}^{f},x_{i}^{v})\}_{i=1}^{N}$, where $x_{i}^{f}$ and $x_{i}^{v}$ are the face and voice examples of the $i_{th}$ instance, respectively. Each pair of an instance $(x_{i}^{f},x_{i}^{v})$  has an associated label $y_{i}\in\{0,1\}$, where $y_{i}=1$ if $x_{i}^{f}$ and $x_{i}^{v}$ belong to the same identity and $y_{i}=0$ if $x_{i}^{f}$ and $x_{i}^{v}$ belong to a different identity.
Both face and voice embeddings typically lie in different representation spaces owing to their different superficial statistics and are mostly unaligned semantically, rendering them incomparable for cross-modal tasks. Cross-modal learning aims at projecting both into a common yet discriminative representation space, where they are sufficiently aligned and instances from the same identity are nearby while from a different are far apart.

\noindent \textbf{Two-stream pipeline.} We employ a two-stream pipeline \cite{nagrani2018learnable} to obtain the respective feature embeddings of both face and voice inputs. The first stream corresponds to a pre-trained convolutional neural network (CNN) on image modality. We take the penultimate layer's output, denoted as $\mathbf{b}_i$, of this CNN as the feature embeddings for an input face image. Likewise, the second stream is a pre-trained audio encoding network that outputs a feature embedding, denoted as $\mathbf{e}_{i}$, for an input audio signal (typically a short-term spectrogram). Existing approaches handling face-voice retrieval \cite{nagrani2018learnable,nawaz2019deep}, mostly resort to triplet and contrastive objectives with carefully crafted negative mining strategies, which significantly increases computational time and are performance-sensitive, to learn a discrminative embedding space. To this end, we introduce a light-weight mechanism that exploits complementary cues from both modality embeddings to form enriched fused embeddings and imposes orthogonal constraints on them for learning discriminative joint face-voice embeddings.

\subsection{Learning Discriminative Joint Embedding}
\label{subsection:Learning_Discriminative_Joint_Embeddings}

In this section, we first describe extracting complementary cues, via multimodal fusion, from both face and voice embeddings obtained through their respective pre-trained networks. We then discuss clustering fused embeddings belonging to the same identity and pushing away the ones with different identity via orthogonality constraints.

Prior to multimodality fusion, we project the face embeddings $\mathbf{b}_i \in \mathbb{R}^{F}$ to a new d-dimensional embedding space $ \mathbf{u}_i \in \mathbb{R}^{d}$ with a fully-connected layer. Similarly, we project the voice embedding $\mathbf{e}_i \in \mathbb{R}^{V}$ to a similar d-dimensional embedding space $ \mathbf{v}_{i} \in \mathbb{R}^{d}$ with another fully-connected layer. We then L2 normalize both $\mathbf{u}_i$ and $\mathbf{v}_i$ which can now be fused to get $\mathbf{l}_{i}$, using the procedure described next.

\noindent \textbf{Multimodal fusion.} We propose to extract complementary features from both modalities, some of which could be related to age, gender and nationality, to form an enriched unified feature representation which is crucial towards learning a discriminative joint embedding space.
Inspired by \cite{arevalo2017gated, chen2020multi}, we employ an attention mechanism to first compute the attention scores (affinity) between the embeddings of two modalities and then fuse these individual modality embeddings after recalibrating them with the attention scores (see Fig.~\ref{fig:overall_framework_fusion}). 
%
We compute attention scores $\mathbf{k}$ between $\mathbf{u}_i$ and $\mathbf{v}_i$  as:
\begin{equation}
    \mathbf{k} = \sigma(F_{att}([\mathbf{u}_i,\mathbf{v}_i]),
\end{equation}

\noindent where $\sigma$ is a sigmoid operator, and $F_{att}$ are the attention layers. Finally, we fuse $\mathbf{u}_i$ and $\mathbf{v}_i$ after modulating them with the attention scores $\mathbf{k}$ to obtain the fused embeddings $\mathbf{l}_{i}$ as:

\begin{equation}
   \mathbf{l}_{i}  = \mathbf{k} \odot tanh(\mathbf{u}_i) + (1 - \mathbf{k}) \odot tanh(\mathbf{v}_i),
\end{equation}

\noindent where $\odot$ is element-wise multiplication.


\noindent \textbf{Supervision via orthogonality constraints.} We want the fused embeddings to encapsulate the semantics of the identity. In other words, these embeddings should be able to predict the identity labels with good accuracy. This is possible if the instances belonging to the same identity are placed nearby whereas the ones with different identity labels are far away. A popular choice to achieve this is softmax cross entropy (CE) loss, which also allows stable and efficient training. Specifically, we use an identity linear classifier with weights denoted as $\mathbf{W}=[\mathbf{w}_{1},\mathbf{w}_{2},...,\mathbf{w}_{C}] \in \mathbb{R}^{d \times C}$ to compute the logits corresponding to $\mathbf{l}_{i}$. Where $d$ is the dimensionality of embeddings and $C$ is the number of identities. Now, identity classification loss with fused embeddings is computed as:

\begin{equation}
    \mathcal{L}_{CE} = -log \frac{exp(\mathbf{l}_{i}^{T}\mathbf{w}_{y_{i}})}{\sum_{j=1}^{C}exp(\mathbf{l}_{i}^{T}\mathbf{w}_{j})}
\end{equation}

Since softmax CE loss does not enforce margins between pair of identities, it is prone to constructing differently-sized class regions which affects identity separability \cite{deng2019arcface,hayat2019gaussian}. 
Some works attempt to include margin between classes in the Euclidean space \cite{wen2016discriminative,calefati2018git}, which is not well synergized with the CE loss as it achieves separation in the angular domain.
Therefore, we propose to impose orthogonality constraints on the fused embeddings to explicitly minimize intra-identity separation while maximizing inter-identity separability \cite{ranasinghe2021orthogonal}. These constraints complement better with the innate angular characteristic of CE loss. Further, since they directly operate on mini-batches, they show greater training efficiency compared to the complex negative mining procedures required in contrastive and triplet loss formulations \cite{nagrani2018learnable, schroff2015facenet} (sec.~\ref{section:experi}).
Formally, the constraints enforce fused embeddings of different identities to be orthogonal and the fused embeddings with same identity to be similar:

\begin{equation}
  \mathcal{L}_{OC}  =  1 - \sum_{i,j \in B, y_{i}=y_{j}} \langle \mathbf{l}_{i},\mathbf{l}_{j}\rangle + \displaystyle\left\lvert \sum_{i,j \in B, y_{i} \neq y_{k}} \langle \mathbf{l}_{i},\mathbf{l}_{k}\rangle\right\rvert,
    \label{Eq:OC}
\end{equation}

\noindent where $ \langle.,.\rangle$ is the cosine similarity operator, and $B$ represents the mini-batch size. The first term in Eq.~\ref{Eq:OC} ensures intra-identity compactness, while the second term enforces inter-identity separation. Note, the cosine similarity involves the normalization of fused embeddings, thereby projecting them to a unit hyper-sphere:

\begin{equation}
    \langle \mathbf{l}_{i},\mathbf{l}_{j}\rangle = \frac{\mathbf{l}_{i}.\mathbf{l}_{j}}{\lVert \mathbf{l}_{i} \rVert_{2}. \lVert \mathbf{l}_{j} \rVert_{2}}.
\end{equation}

\noindent \textbf{Overall Training Objective.} To train the proposed framework, we minimize the joint loss formulation, comprising of $\mathcal{L}_{CE}$ and $\mathcal{L}_{OC}$ as:


\begin{equation}
\label{eq:floss}
    \mathcal{L} = \mathcal{L}_{CE} + \alpha \mathcal{L}_{OC},
\end{equation}

\noindent where $\alpha$ balances the contribution of two terms in $\mathcal{L}$. We empirically set $\alpha$ to 1.0 based on validation set performance. It is important to mention that CE loss operates in logit space and orthogonal constraints are imposed in the embedding space, however, both of them synergizes well with each other owing to their common angular domain characteristic.

\begin{figure*}
\centering
\includegraphics[scale=0.55]{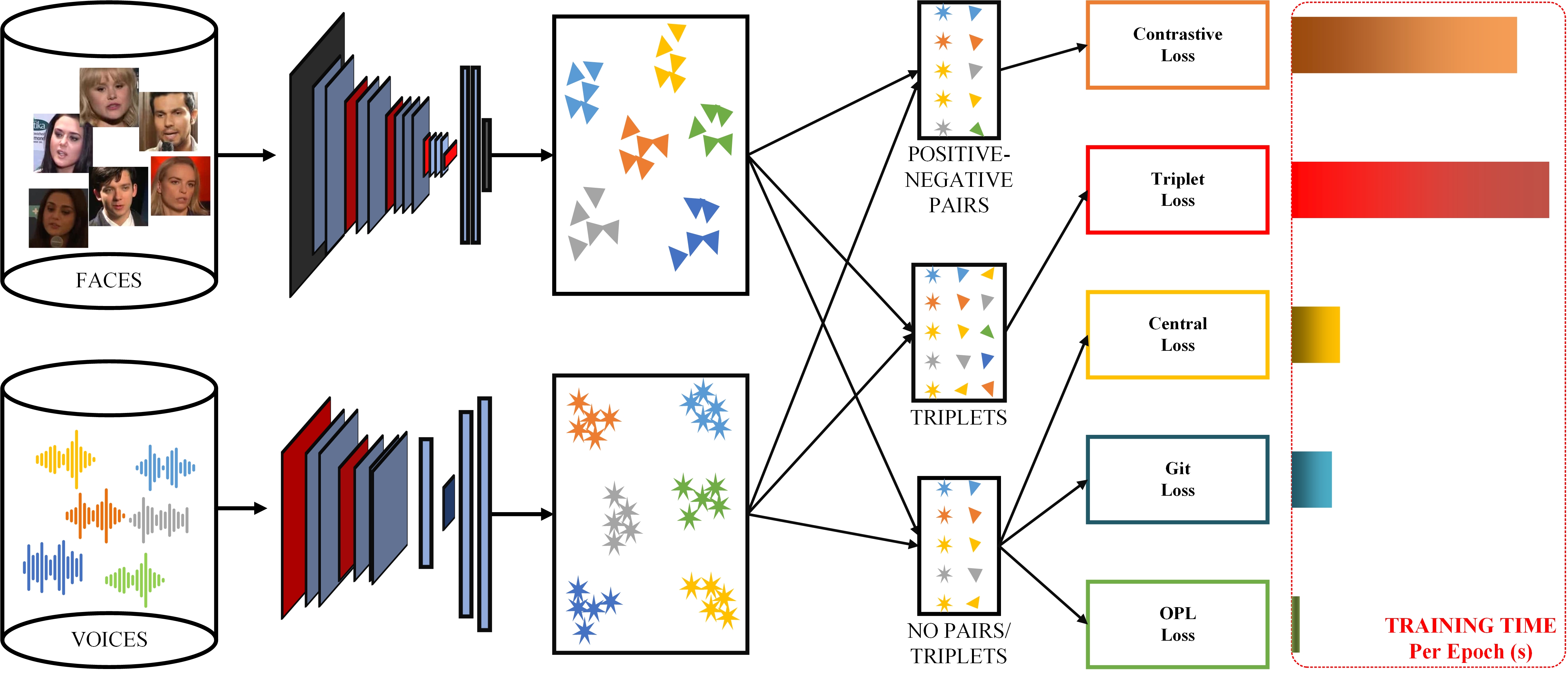}
   \caption{Multimodal frameworks based on two-stream network to embed face and voice modalities in the latent space.  We compare our loss formulation with various others typically adopted in existing face-voice association methods, including Contrastive~\cite{nagrani2018learnable}, Triplet~\cite{nagrani2018seeing}, Center~\cite{nawaz2019deep} and Git~\cite{calefati2018git}, under the same feature embeddings, network architecture, and evaluation protocol with an aim to provide a fair comparison. 
   } 
\label{fig:comparison}
\end{figure*}

\subsection{Existing Loss Formulations in F-V Methods}
In this section, we formally overview several existing loss formulations typically employed in existing face-voice association methods with an aim to provide a fair comparison with ours, see Fig.~\ref{fig:comparison}.

\noindent \textbf{Center Loss.}  It simultaneously learns class centers from features in a mini-batch and penalizes the distance between each class center and corresponding features~\cite{wang2016learning}. Recently, Nawaz et al.~\cite{nawaz2019deep} introduces a single stream architecture to extract audio-visual information to bridge the gap between them, leveraging center loss.  The loss is formulated as follow:

\begin{equation}
\label{eq:closs}
\mathcal{L}_{C}=\frac{1}{2} \sum_{i=1}^{b}\left\|\mathbf{I}_{i}-\mathbf{c}_{y_{i}}\right\|_{2}^{2}
\end{equation}

The $\mathbf{c}_{y_{i}}$ represents the $\mathbf{y_{i}}$th class center of features. Wen et al.~\cite{wen2016discriminative} observed that the center loss is very small which may degraded to zeros, thus, it is jointly trained with CE loss as follow:

\begin{equation}
\label{eq:closs_softmax}
    \mathcal{L} = \mathcal{L}_{CE} + \alpha_c \mathcal{L}_{C}
\end{equation}

A scalar value $\alpha_c$ is used for balancing center loss and CE loss.

\noindent \textbf{Git Loss.}  It improves center loss by maximizing the distance between features belonging to different classes (push) while keeping features of the same class compact (pull)~\cite{calefati2018git}. The loss is formulated as follow:

\begin{equation}
\label{eq:git}
\begin{aligned}
\mathcal{L}_{G} &=  \sum_{i,j=1,i\neq j}^b  \frac{1}{1 + \left\Vert \mathbf{I}_{i} - \mathbf{c}_{y_j} \right\Vert^2_2}
\end{aligned}
\end{equation}

Git loss is trained jointly trained with center loss and CE loss as follow:

\begin{equation}
\label{eq:closs_git_softmax}
    \mathcal{L} = \mathcal{L}_{CE} + \alpha_c \mathcal{L}_{C}  + \alpha_g \mathcal{L}_{G}
\end{equation}

Scalar values $\alpha_c$ and $\alpha_g$ are used for balancing center loss, git loss and CE loss.

\noindent \textbf{Contrastive Loss~\cite{nagrani2018learnable}.} It construct positive and negative pairs to learn joint embedding. A set $\rho$ is formulated comprising of training pairs $\{(\mathbf{u}_i, \mathbf{v}_i)\}$ using d-dimensional face  ($\mathbf{u}_i$)  and voice ($\mathbf{v}_i$) embeddings. Each pair has an associated label $y_{i}\in\{0,1\}$, where $y_{i}=1$ if $\mathbf{u}_i$ and $\mathbf{v}_i$ belong to the same class and $y_{i}=0$ if $\mathbf{u}_i$ and $\mathbf{v}_i$ belong to different class.

\noindent \textbf{Triplet Loss~\cite{nagrani2018seeing}.} It encourages that dissimilar pairs be distant from  similar pairs by a certain margin value by formulating a triplet. A triplet consists of  an anchor, positive and negative sample to formulate a triplet. 
Given $d-$dimensional face embedding $\textbf{u}_{i}$ and voice embeddings $\textbf{v}_{i}$, we take a voice anchor $\textbf{v}_{i}^a$ and choose a positive face  $\textbf{u}_{i}^{p}$ and negative face $\textbf{u}_{i}^{n}$ constructing a triplet~\{$\textbf{v}_{i}^{a}$, $\textbf{u}_{i}^{p}$, $\textbf{u}_{i}^{n}$\} such that:

\begin{equation}
    \||\mathbf{v}_{i}^{a}-\mathbf{u}_{i}^{p}||_{2}^{2} < ||\mathbf{v}_{i}^{a}-\mathbf{u}_{i}^{n}||_{2}^{2}~~\forall(v_{i}^{a}, u_{i}^{p}, u_{i}^{n})\in\rho
\end{equation}

where $\rho$ is the set of all triplets. In this work, we have used negative hard mining to generate triplets~\cite{schroff2015facenet}.  The loss function can now be defined as:
\begin{equation}
    \mathcal{L}_{triplet} = \sum_{i=1}^{N} [||\mathbf{v}_{i}^{a}-\mathbf{u}_{i}^{p}||_{2}^{2}-||\mathbf{v}_{i}^{a}-\mathbf{v}_{i}^{n}||_{2}^{2}+m]
\end{equation}
where $m$ is margin factor which helps keeping a distance between anchor and positive and negative samples. 
The loss function  minimizes the distance between the anchor voice and positive face i.e. push $||\textbf{v}_{i}^{a}-\textbf{u}_i^{p}||$ and penalize $||\textbf{v}_{i}^{a}-\textbf{u}_{i}^{n}||$ to be greater than $||\textbf{v}_{i}^{a}-\textbf{u}_i^{p}|| + m$. Intuitively, this helps in producing closely packed clusters of samples from same class while simultaneously pushing apart clusters of other classes.

\section{Experiments}
\label{section:experi}
We perform various cross-modal experiments on \textit{verification} and \textit{matching} tasks to evaluate the effectiveness of our method and provide ablation study with analysis to show the impact of different components in our method. 
In addition, we analyze the impact of multiple languages on face-voice association.


\subsection{Experimental Setup}
\label{subsection:Experimental_Setup}

\noindent \textbf{Training Details and Datasets.} We train our method on Quadro P$5000$ GPU for $50$ epochs using a batch-size of $128$ using Adam optimizer with exponentially decaying learning rate (initialised to \(10^{-5}\)). We extract face and voice embeddings from VGGFace~\cite{parkhi2015deep} and Utterance Level Aggregation~\cite{xie2019utterance}. Note that, we only backprop. through FOP module while the weights of face and voice subnetworks remains unchanged.

Nagrani et. al~\cite{nagrani2017voxceleb} introduced a large-scale dataset of audio-visual human speech videos extracted `in the wild' from YouTube.
Learnable Pins~\cite{nagrani2018learnable} introduced two train/test splits out of this dataset to perform various cross-modal verification and matching tasks.
The first split consists of disjoint videos from the same set of speakers while the second split contains disjoint identities.
We train the model using these two training sets, allowing us to evaluate on both test sets, the first one for \textit{seen-heard} identities, and the second for \textit{unseen-unheard} identities.
Note that we followed the same train, validation and test split configurations as used in~\cite{nagrani2018learnable} for fair comparisons.

Recently Nawaz et al.~\cite{nawaz2021cross} curated MAV-Celeb dataset to study the impact of language on face-voice association task; it consists of video and audio recordings of celebrities speaking more than one language. The dataset contains two splits English–Urdu (EU) and English–Hindi (EH) to analyze performance measure across multiple languages, leveraging \textit{unseen-unheard} and completely \textit{unheard} evaluation protocol. In this work, we only used EU split in our experiments.

\noindent \textbf{Experimental Setup.} 
We perform our experiments on following two tasks: The first task is to perform \textit{cross-modal verification} where the goal is to verify if an audio segment and a face image belong to the same identity. Two inputs are considered i.e. face and voice and verification between the two depends upon a threshold on the similarity value. The threshold can be adjusted in accordance to wrong rejections of true match and/or wrong acceptance of false match. We report results on standard verification metrics i.e. ROC curve (AUC) and equal error rate (EER).
The second task consists of \textit{cross-modal matching} where the goal is to match the input modality to the varying gallery size $n_c$  which consists of the other modality. In our experiments, we increase $n_c$ to analyze its impact on the performance. For example, in $1:2$ matching task, we are given a modality at input, e.g. face, and the gallery consists of two inputs from other modality, e.g. audio. One of them contains a true match while the other serves as an imposter. We employ matching metric i.e. accuracy to report performance. We perform this task in five settings where in each setting the $n_c$ is increased as $2,4,6,8,10$.

\subsection{Results}

\begin{table*}
\parbox{.45\linewidth}{
\caption{Cross-modal verification results for our (joint) loss and other losses under two configurations and two error metrics.}
\resizebox{1.11\linewidth}{!}{
\begin{tabular}{|lcc|cc|c|}
\hline
Method  & EER & AUC & EER & AUC\\
\hline
 & \multicolumn{2}{c|}{Seen-Heard} & \multicolumn{2}{c|}{Unseen-Unheard}\\
\hline\hline
CE Loss                                                & 21.8 & 86.6  & 26.8 & 81.7\\
Center Loss~\cite{wen2016discriminative,nawaz2019deep} & 19.8 & 88.6  & 29.7 & 77.5\\
Git Loss~\cite{calefati2018git}                        & 19.6 & 88.9  & 29.5 & 77.8\\
Contrastive Loss~\cite{nagrani2018learnable}           & 23.4 & 84.7  & 29.1 & 79.5\\
Triplet Loss~\cite{schroff2015facenet}                 & 20.7 & 88.0   & 27.1 & 81.4\\ 
Ours                                                   & \textbf{19.3 }& \textbf{89.3}  & \textbf{24.9} & \textbf{83.5}\\
\hline
\end{tabular}
\label{tab:result-base}
}
}
\hfill
\parbox{.45\linewidth}{
\caption{Theoretical/empirical training complexity of our (joint) loss and others. $n$: \# training instances in a modality, and $B$ is mini-batch size.}
\resizebox{1.0\linewidth}{!}{
\begin{tabular}{|lcc|cc|c|}
\hline
Method  & Empirical & Theoretical\\
\hline
 & Time (s)  & Worst Case\\

\hline\hline
CE Loss   & .02  & $\mathcal{O} (n)$ \\
Center Loss~\cite{wen2016discriminative,nawaz2019deep}  & 6.8 & $\mathcal{O} (n+\frac{n^2}{B})$\\
Git Loss~\cite{calefati2018git}                        &  6.2 & $\mathcal{O} (n+\frac{n^2}{B})$\\
Contrastive Loss~\cite{nagrani2018learnable}           & 568.2  & $\mathcal{O} (n^2)$\\
Triplet Loss~\cite{schroff2015facenet}                 & 619.7 & $\mathcal{O} (n^3)$\\ 
Ours                                                   & 0.7  & $\mathcal{O} (n)$ \\
\hline

\end{tabular}
\label{tab:runtime-result-base}
}
}
\end{table*}

\noindent \textbf{Comparison with other F-V losses.} We compare our proposed FOP against various losses typically employed in F-V association methods. 
The first is \textit{center loss}~\cite{wen2016discriminative}, and is adapted by ~\cite{nawaz2019deep} in a single stream network to learn F-V association for cross-modal verification and matching tasks.
The second is \textit{Git loss} ~\cite{calefati2018git}, which in addition to center loss also maximizes the intra-identity distances of embeddings with other centroids and has shown effectiveness for verification and matching tasks. 
In implementation, we simply replaced our orthogonal constraints loss formulation ($\mathcal{L}_{OC}$) with center loss and then Git loss while keeping all other settings unaltered.
%
The third one is \textit{Contrastive Loss} that is adapted by ~\cite{nagrani2018learnable} to associate face and voice of a person in a cross-modal verification task. 
The last is \textit{Triplet Loss}, that was leveraged by ~\cite{nagrani2018seeing} in their face-voice association framework.
In implementation, we remove our FOP module in the overall architecture and plug first contrastive and then triplet loss with hard negative mining strategy~\cite{schroff2015facenet} while keeping the rest of settings fixed. 


Table~\ref{tab:result-base} shows the comparison of our supervision formulation with center and Git losses, and our FOP mechanism with the contrastive and triplet losses on cross-modal verification task with \textit{seen-heard} and \textit{unseen-unheard} configurations. We observe that our proposal performs better than other loss formulations across all configurations and both error metrics. We can attribute this improved performance across full spectrum to two possible reasons. First, our orthogonality constraints don't depend on some margin hyperparameter which is certainly difficult to tune with large number of instances and identities (as in VoxCeleb$1$). Second, as opposed to centroid based losses e.g., center or Git, these orthogonality constraints complement well with the innate angular characteristic of softmax CE loss.
We compare both the theoretical and empirical run-time training complexity of our (joint) loss formulation and various others commonly employed in existing F-V methods (Table~\ref{tab:runtime-result-base}). The empirical run-time is the time to complete one epoch. Our loss formulation is superior than all others listed in terms of both theoretical and empirical training efficiency.
To further validate the effectiveness of our proposed FOP mechanism, we examine the effect of Gender (G), Nationality (N), Age (A) and its combination (GNA) separately, which influence both face and voice verification (Table~\ref{tab:results-demographic}). FOP achieves consistently better performance on  G, N, A and the combination (GNA) in both \textit{seen-heard} and \textit{unseen-unheard} configurations than other loss formulations. 





\begin{table*}[!htp]
\caption{Cross-modal biometrics results under varying demographics for \textit{seen-heard} and \textit{unseen-unheard} configurations.}
\centering
\begin{tabular}{|llcccc|lcccc|}
\hline
Demographic   &Random & G & N & A & GNA &Random & G & N & A & GNA \\
\hline
 & \multicolumn{5}{c|}{Seen-Heard}  & \multicolumn{5}{c|}{Unseen-Unheard}\\
\hline\hline
CE & 86.6 & 78.0 & 85.0 & 86.3 & 77.3 & 81.7 & 65.9 & 53.6 & 76.0 & 52.8 \\
Center & 88.6  & 79.2   & 87.0  &  88.2  & 78.1  & 77.5  & 62.4  & 51.7  &  72.5  & \textbf{54.2} \\
Git  & 88.9  & \textbf{79.7}   & 87.4  &  \textbf{88.6}  & \textbf{78.5} & 77.9  & 62.6  & 51.8  &  72.8  & \textbf{54.2}   \\
Contrastive  & 84.7  & 69.7     & 83.7     &   84.5    & 69.2   & 79.5  & 61.0     & 53.5     &   74.7    & 51.8  \\
Triplet  & 88.0  & 76.3   & 86.7  &  87.6  & 75.6 & 81.7  & 65.5  & 53.4  & 76.3   & 52.2  \\
Ours                                                   & \textbf{89.3}  & 76.7  & \textbf{87.9}   &   \textbf{88.6}   &76.6  & \textbf{83.5}  & \textbf{68.8}  & \textbf{54.9}  & \textbf{78.1}   & \textbf{54.2}  \\

\hline
\end{tabular}
\label{tab:results-demographic}
\end{table*}

\begin{figure}[!htp]
\centering
\includegraphics[scale=0.5]{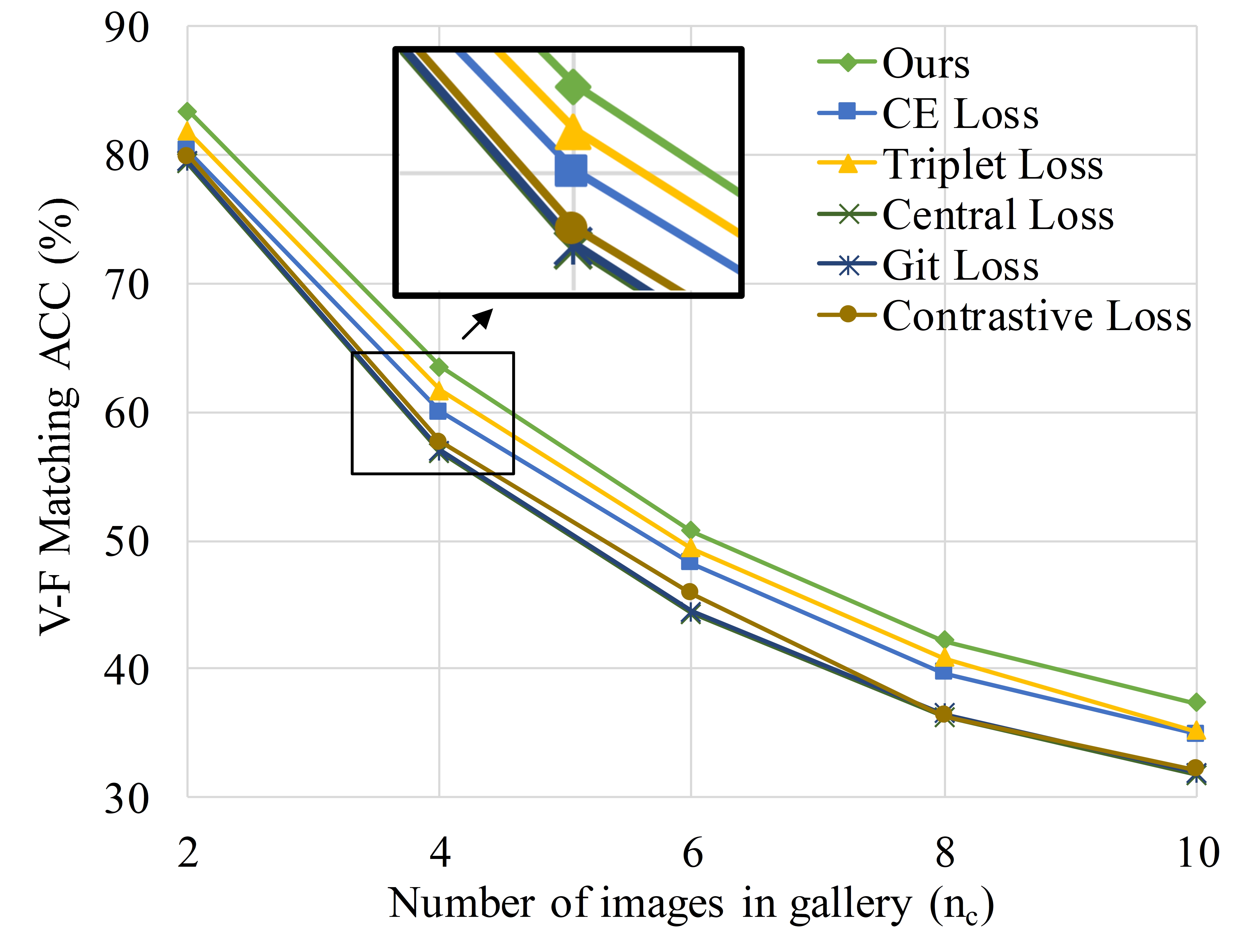}

   \caption{\small Cross-modal matching results with varying gallery size for our proposed FOP and different loss formulations employed by the existing face-voice association methods.}
\label{fig:result-nway}
\end{figure}

\begin{figure*}[t!]
\begin{center}
\includegraphics{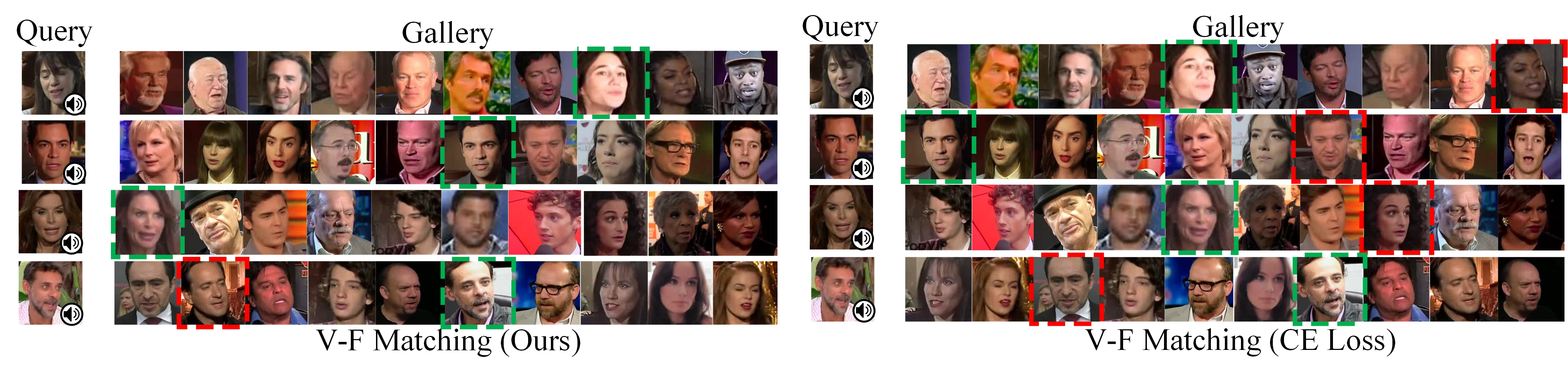}
\end{center} 
   \caption{Qualitative evaluation of V-F matching task: We evaluate the performance of our method by examining the failure cases of CE Loss. A voice query is shown on the left, and a gallery of 10 template images are shown on the right. There are three successful predictions by our proposed method (highlighted in green). On the last row the wrong prediction is highlighted in red whereas ground truth is highlighted in green. It is clear that the embedding produced with FOP is discriminative in contrast to the embedding produced by the traditional CE loss.  Best viewed zoomed in and in colour.}
\label{fig:quali}
\end{figure*}

Furthermore, we compare our FOP against aforementioned loss functions on a cross-modal matching task, $1:n_c$ with $n_c=2,4,6,8,10$ in Fig.~\ref{fig:result-nway}. %
In this setting, the input is voice while the matching gallery consists of faces. For example, consider the case where the input is voice and the task is $1:2$ matching, we find that out the entry in gallery which best matches the input. We see that our proposed FOP outperforms the counterpart loss formulations for all values of $n_c$. Note, this is a relatively challenging task as the common trend for all competitors is that upon increasing $n_c$ the matching performance deteriorates. In addition to the numerical results, we present qualitative results in Fig.~\ref{fig:quali}.

\begin{figure}
\begin{center}
\includegraphics[scale=0.4]{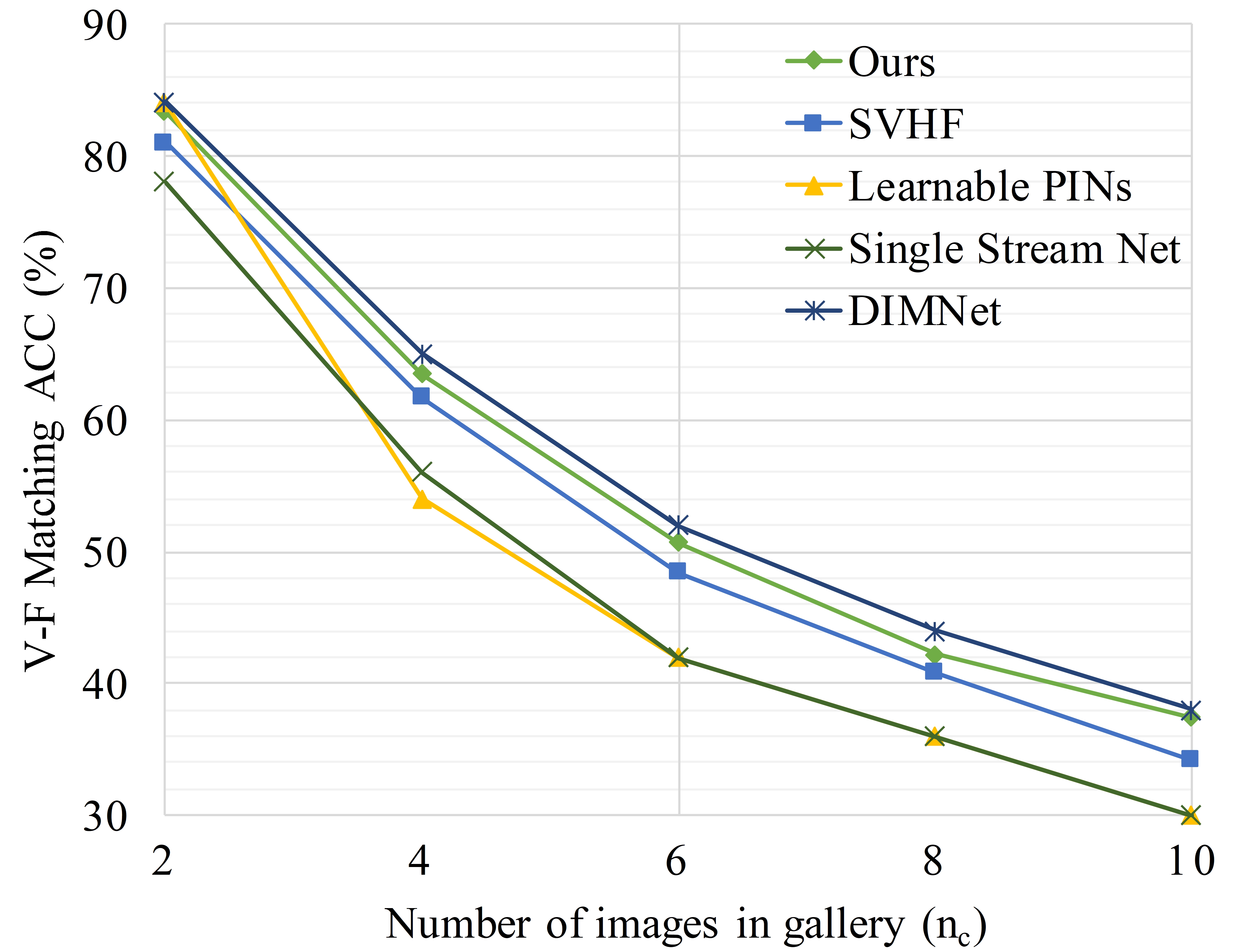}
\end{center} \vspace{-1em}
   \caption{\small Cross-modal matching results of our method and existing state-of-the-art methods with varying gallery size.}
\label{fig:nway-sota}
\end{figure}

\noindent \textbf{Comparison with state-of-the-art.} Table~\ref{tab:sota} compares our method against existing state-of-the-art works (DIMNet~\cite{wen2018disjoint}, Learnable Pins~\cite{nagrani2018learnable}, MAV-Celeb~\cite{nawaz2021cross}, Single Stream Network~\cite{nawaz2019deep}, Multi-view Approach~\cite{sari2021multi}, Adversarial-Metric Learning~\cite{zheng2021adversarial}, Disentangled Representation Learning~\cite{ning2021disentangled}, Voice-face Discriminative Network~\cite{tao20b_interspeech}). Our method shows the best performance on \textit{unseen-unheard} protocol, however, it achieves the second best performance on \textit{seen-heard} protocol. Adversarial-Metric Learning~\cite{nawaz2019deep} records top performance on \textit{seen-heard} configuration.
We further show a comparison of our method with existing state-of-the-art methods on cross-modal matching (Fig.~\ref{fig:nway-sota}). In particular, we perform the $1:n_c$  matching tasks, where $n_c=2,4,6,8,10$ and report the results. Our method outperforms~\cite{nagrani2018learnable} while achieves competitive performance against DIMNet~\cite{wen2018disjoint}. Note that, in addition to identity, DIMNet leverages two additional source of supervisions from gender and nationality, which are not always readily available.

\begin{table*}[!htp]
\caption{ Cross-modal verification results on \textit{seen-heard} and \textit{unseen-unheard} configurations of our method and existing state-of-the-art methods.}
\centering

\begin{tabular}{|lccc|cc|}
\hline
Methods  & Dataset & EER & AUC  & EER & AUC \\
\hline
& & \multicolumn{2}{c|}{Seen-Heard} & \multicolumn{2}{c|}{Unseen-Unheard}\\
\hline\hline
Voice-face Discriminative Network~\cite{tao20b_interspeech}      &   VoxCeleb2      &  -    & -       &  22.5       &  85.4     \\
\hline
DIMNet~\cite{wen2018disjoint}                                    &   VoxCeleb1      & -    & -      & 24.9          & -          \\
Learnable Pins~\cite{nagrani2018learnable}                       &   VoxCeleb1      & 21.4 & 87.0   & 29.6          & 78.5        \\
MAV-Celeb~\cite{nawaz2021cross}                                  &   VoxCeleb1      & -    & -      & 29.0          & 78.9         \\
Single Stream Network~\cite{nawaz2019deep}                       &   VoxCeleb1      & \textbf{17.2} & 91.1          &  29.5 & 78.8  \\
Multi-view Approach~\cite{sari2021multi}                         &   VoxCeleb1      & -    & -      &  28.0         & -               \\
Adversarial-Metric Learning~\cite{zheng2021adversarial}          &   VoxCeleb1      & -    & \textbf{92.3 }  & -             &  80.6            \\
Disentangled Representation Learning~\cite{ning2021disentangled} &   VoxCeleb1      & -    & -      & 25.0          &  \textbf{84.6}             \\
Ours                                                             &   VoxCeleb1      & 19.3 & 89.3   & \textbf{24.9} & 83.5      \\

\hline
\end{tabular}
\label{tab:sota}
\end{table*}

\noindent \textbf{Ablation study and analysis.} Table~\ref{tab:fusion-abl} reports the results when varying the hyperparameter $\alpha$ that is used to balance the contribution of softmax CE and orthogonal constraints based loss in our joint formulation (Eq.~\ref{eq:floss}). Our method performance is mostly robust to the choice of $\alpha$. There is only a 0.8\% drop in EER when varying  $0.5 \leq \alpha \le 2.0$ around the best reported value of 1.0. However, the performance starts to deteriorate beyond 0.5 and above 2.0, where the contribution of softmax CE, which synergizes with orthogonality constraints, is almost negligible. We also experiment with replacing gated multimodal fusion with a much simpler linear  fusion (Table~\ref{tab:linear-vs-gated}). Linear fusion naively sums element-wise the embeddings from two modalities. Results reveal the gated fusion significantly works better than the linear fusion. This highlights the importance of first finding the correlating features between the two modality embeddings, modulating them accordingly, and finally fusing them.


%
%
\begin{table}[!htp]
\caption{Coss-modal  Verification  results  on \textit{unseen-unheard} configurations to illustrate  the effect of varying $\alpha$ value.}
\centering
\begin{tabular}{|l|c|c|c|c|c|c|}
\hline
 \multicolumn{7}{|c|}{$ \mathcal{L}_{CE} + \alpha \mathcal{L}_{OC}$} \\
\hline
 $\alpha$ & 0.0 & 0.1 & 0.5 & 1.0 & 2.0 & 5.0 \\
\hline\hline 
EER  & 26.8 & 26.1 & 25.8 & \textbf{24.9} & 25.9 &26.0 \\
AUC  & 81.7 & 82.4 & 82.8 & \textbf{83.5} & 82.7 & 82.6 \\

\hline
\end{tabular}

\label{tab:fusion-abl}
\end{table}
\begin{table}[!htp]
\caption{\small Coss-modal  Verification  results  on \textit{unseen-unheard} configurations with linear and gated fusion strategies.}
{
\centering
\resizebox{0.30\textwidth}{!}{%
\begin{tabular}{|l|c|c|}
\hline
Fusion Strategy & EER & AUC \\
\hline\hline
Linear Fusion           & 25.6    & 82.7 \\
Gated Fusion & \textbf{24.9}    & \textbf{83.5 } \\
\hline
\end{tabular}}
\vspace{0.5em}

\label{tab:linear-vs-gated}
}
\end{table}

\begin{figure*}[!htp]
\begin{center}
\includegraphics[scale=0.6]{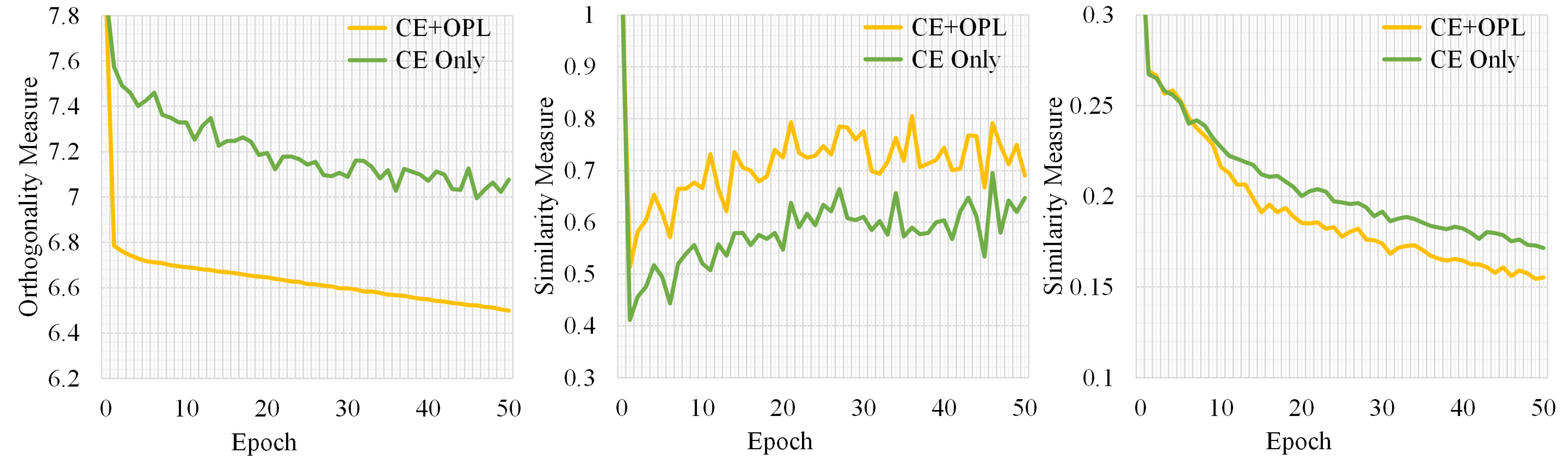}
\end{center} 
 \caption{(a) Feature Orthogonality (↓) (b) Similarity of same class features (↑) (c) Similarity of different class features (↓).  FOP simultaneously achieves higher inter-class similarity and intra-class dissimilarity in comparison with the standard CE. }
\label{fig:decompose}
\end{figure*}

\noindent \textbf{Decomposing Orthogonal Constraints (OC).} Taking a step further, we  break down OC into its sub-components as defined in Eq.~\ref{Eq:OC}. These components are employed to quantify the intra-class compactness and inter-class separation in any given embedding space.
Therefore, it will be interesting to compare the contribution of OC towards inter-class maximisation and intra-class minimization of an embedding space in contrast to the traditional CE based training scenario.
In Fig.~\ref{fig:decompose}, the effect of OC on learning face-voice association is visualized. It is evident that OC has a strong impact on overall feature orthogonality in fused embedding space. Moreover, it helps to increase the similarity index between features of similar classes as compared to only CE while simultaneously enforcing higher intra-class dissimilarity.

\begin{figure}
\begin{center}
\includegraphics[scale=0.5]{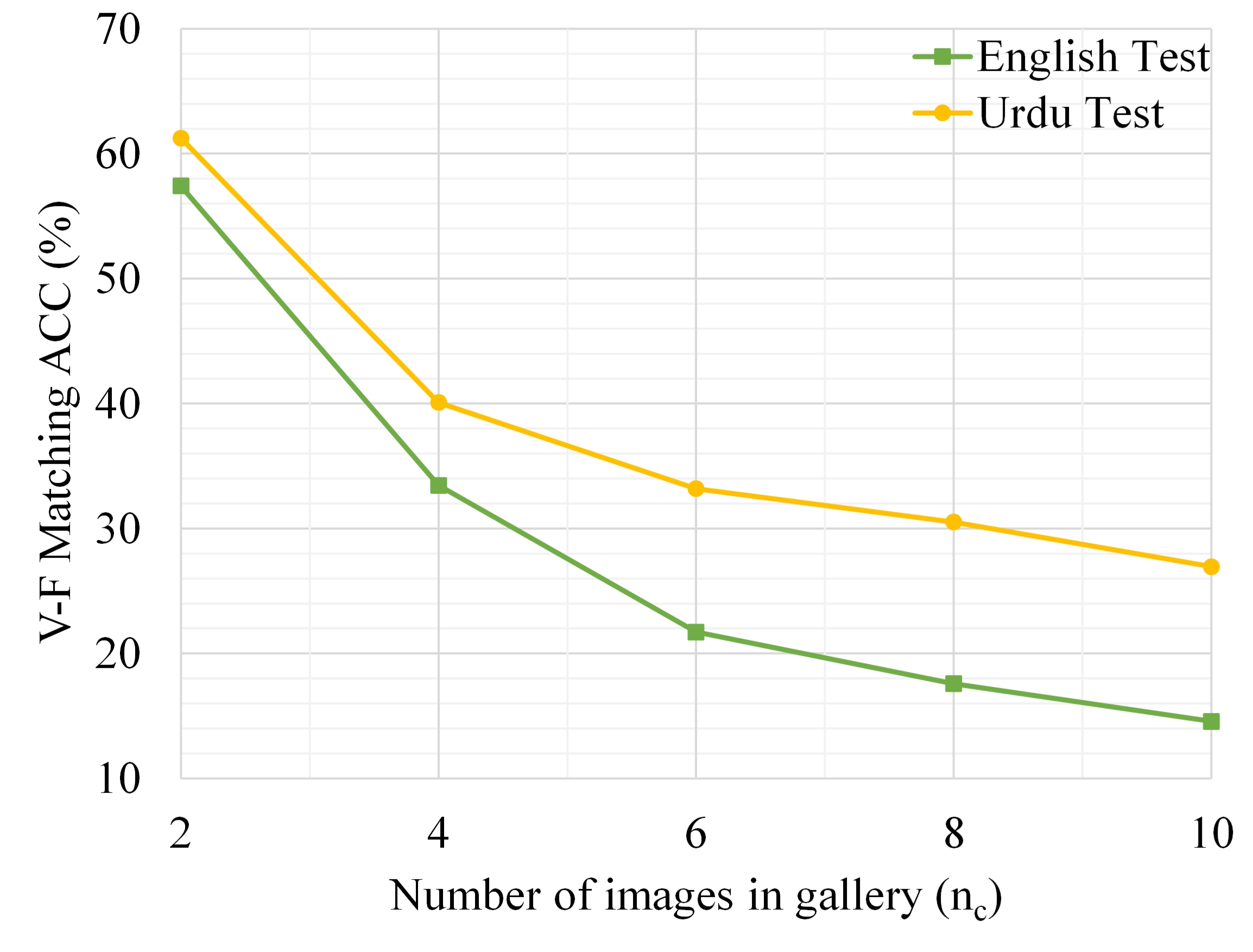}
\end{center}
    \caption{Cross-modal matching results of our method on \textit{heard} language (Urdu) and completely \textit{unheard} language (English).}
\label{fig:nway-urdu-train}
\end{figure}

\begin{figure}[!htp]
\begin{center}
\includegraphics[scale=0.5]{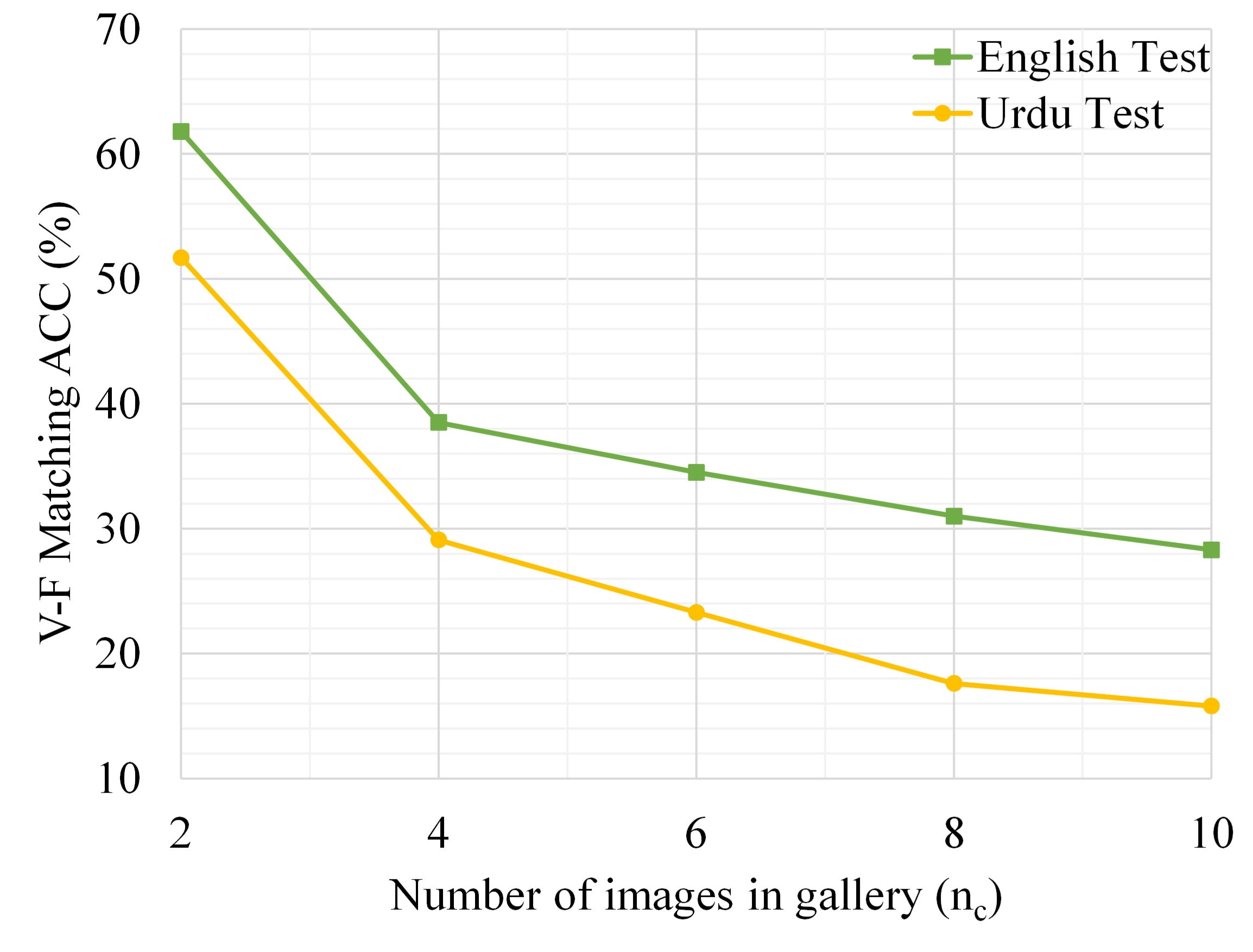}
\end{center}
    \caption{Cross-modal matching results of our method on \textit{heard} language (English) and completely \textit{unheard} language (Urdu).}
\label{fig:nway-eng-train}
\end{figure}

\begin{figure*}[!htp]
\centering  
\subfigure{\includegraphics[width=0.3\linewidth]{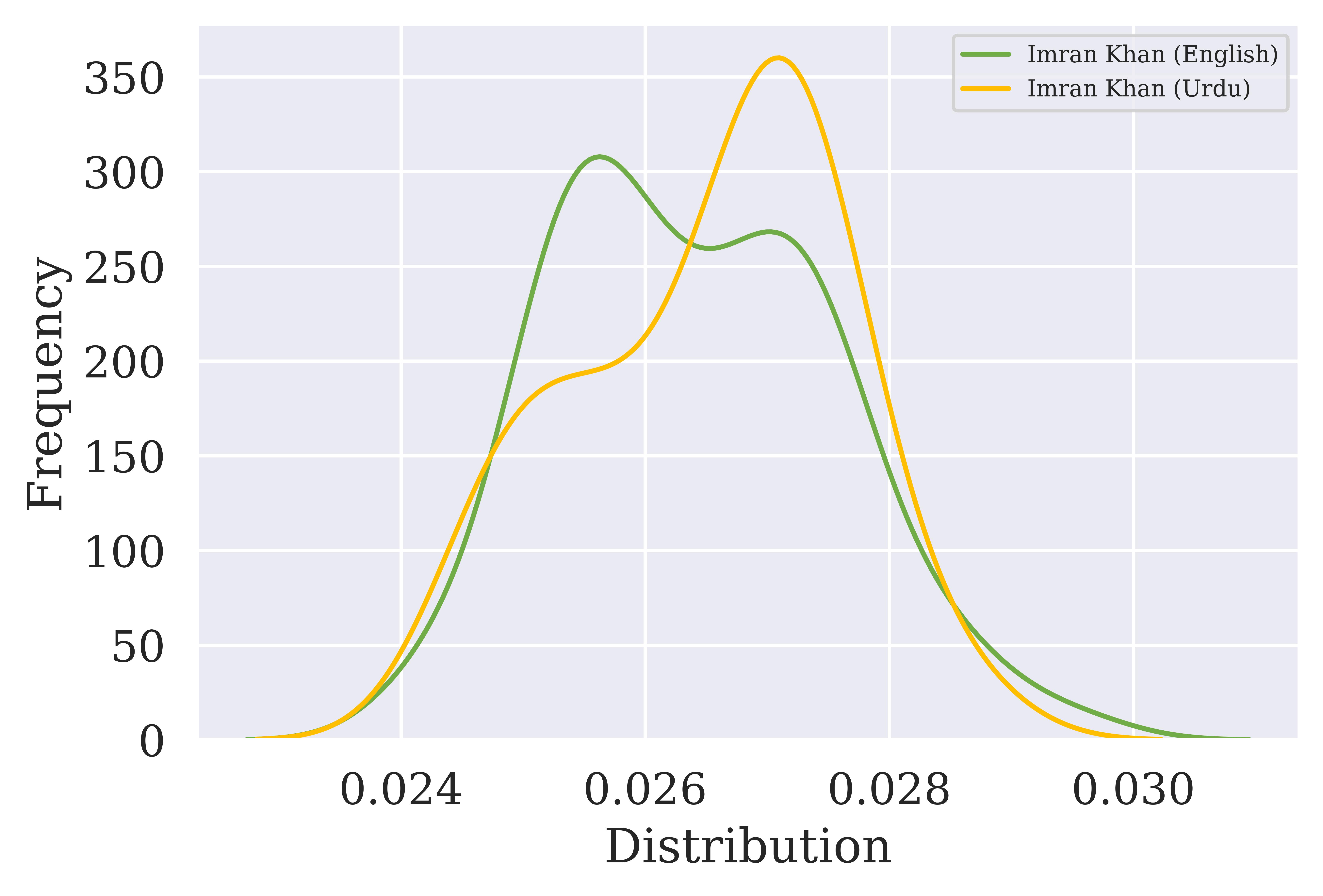}}
\subfigure{\includegraphics[width=0.3\linewidth]{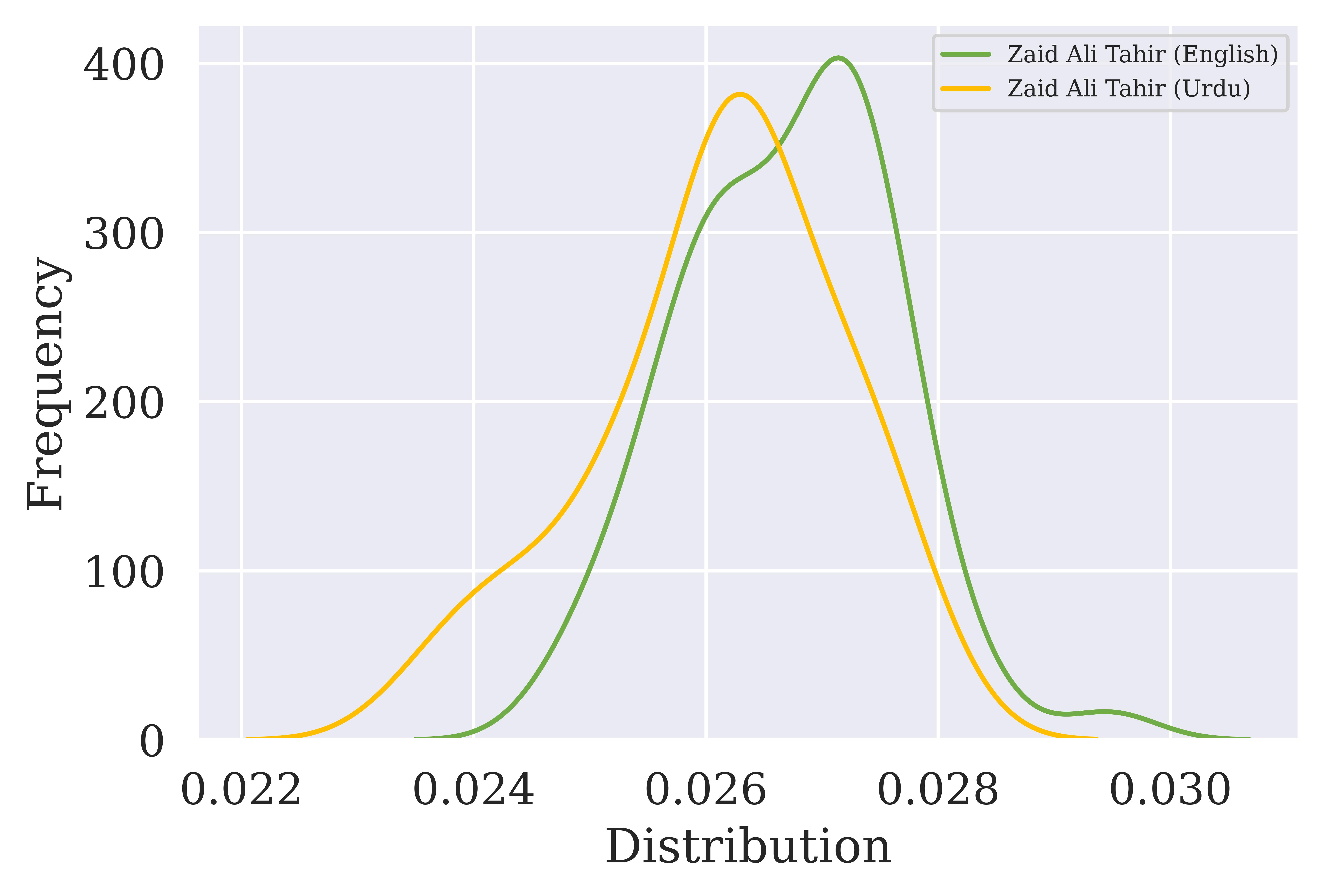}}
\subfigure{\includegraphics[width=0.3\linewidth]{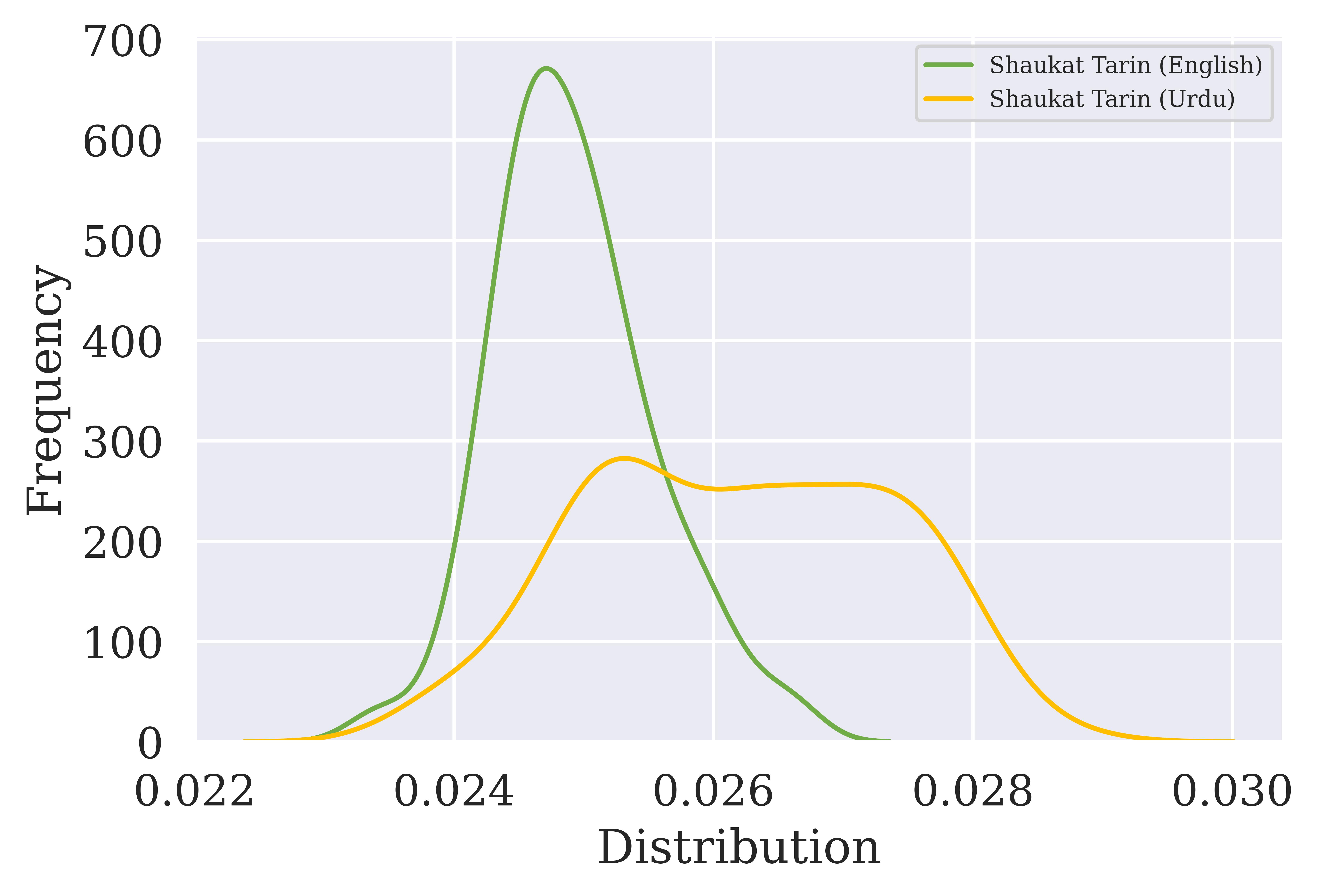}}

\hrulefill

\subfigure{\includegraphics[width=0.3\linewidth]{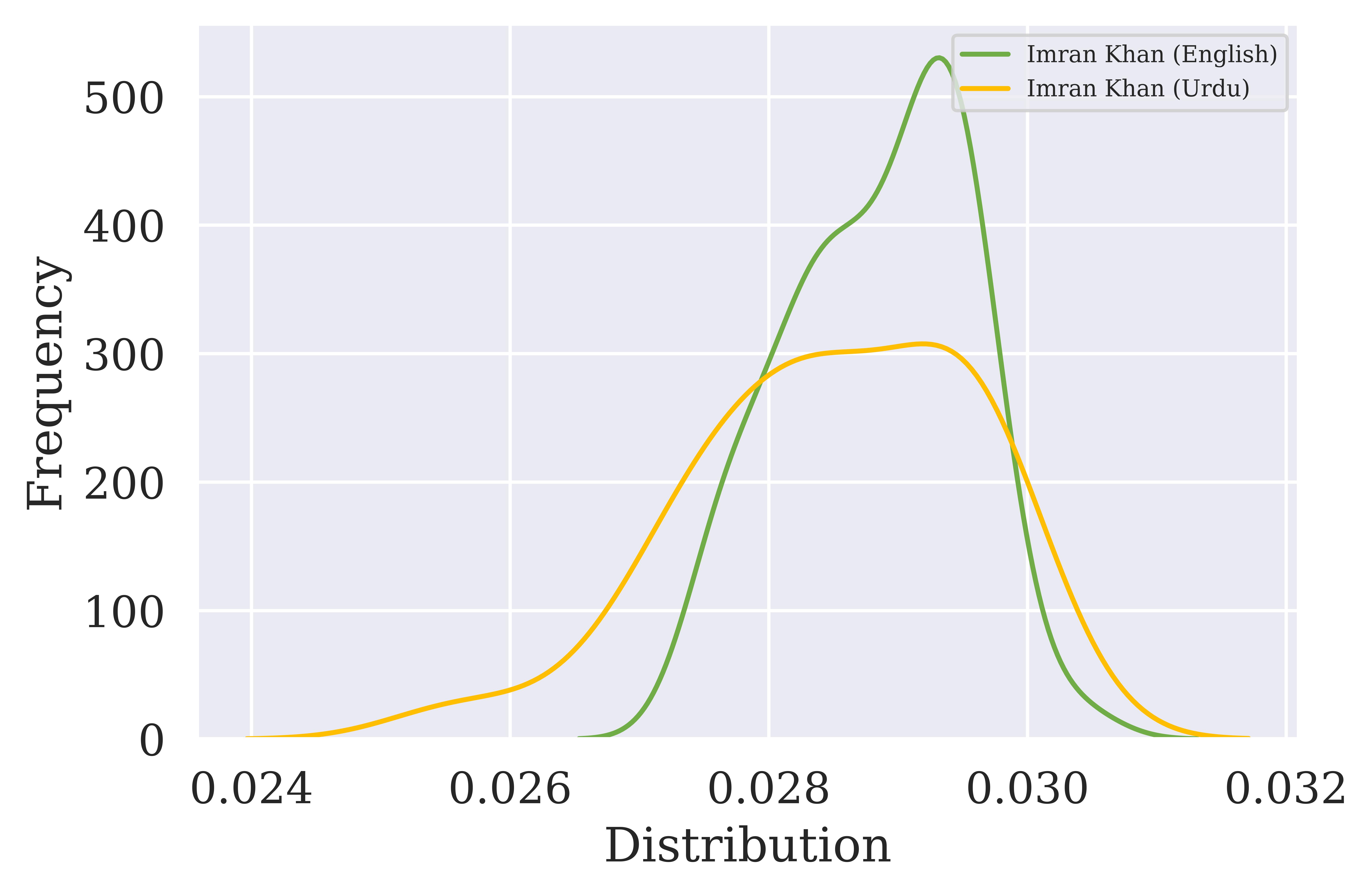}}
\subfigure{\includegraphics[width=0.3\linewidth]{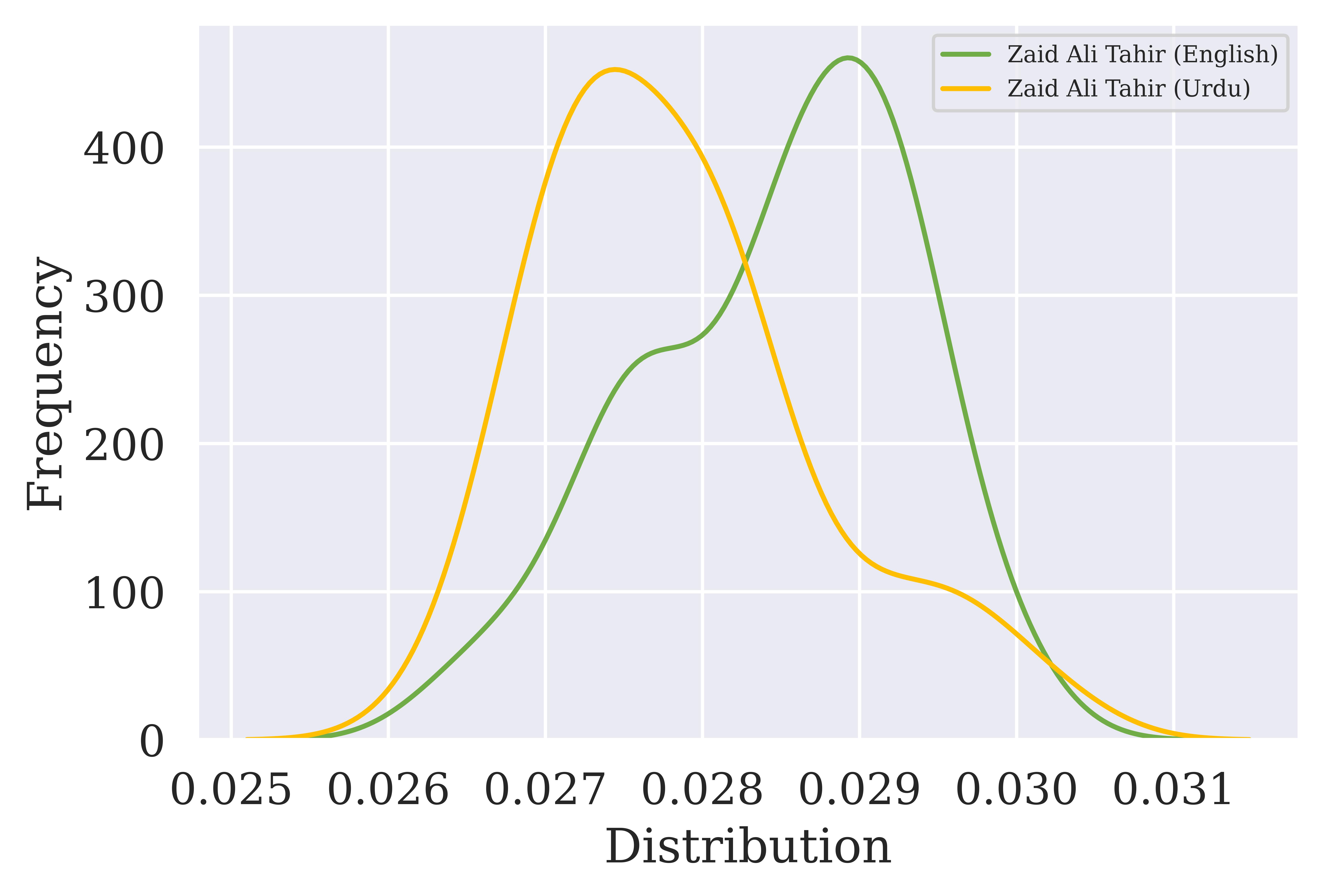}}
\subfigure{\includegraphics[width=0.3\linewidth]{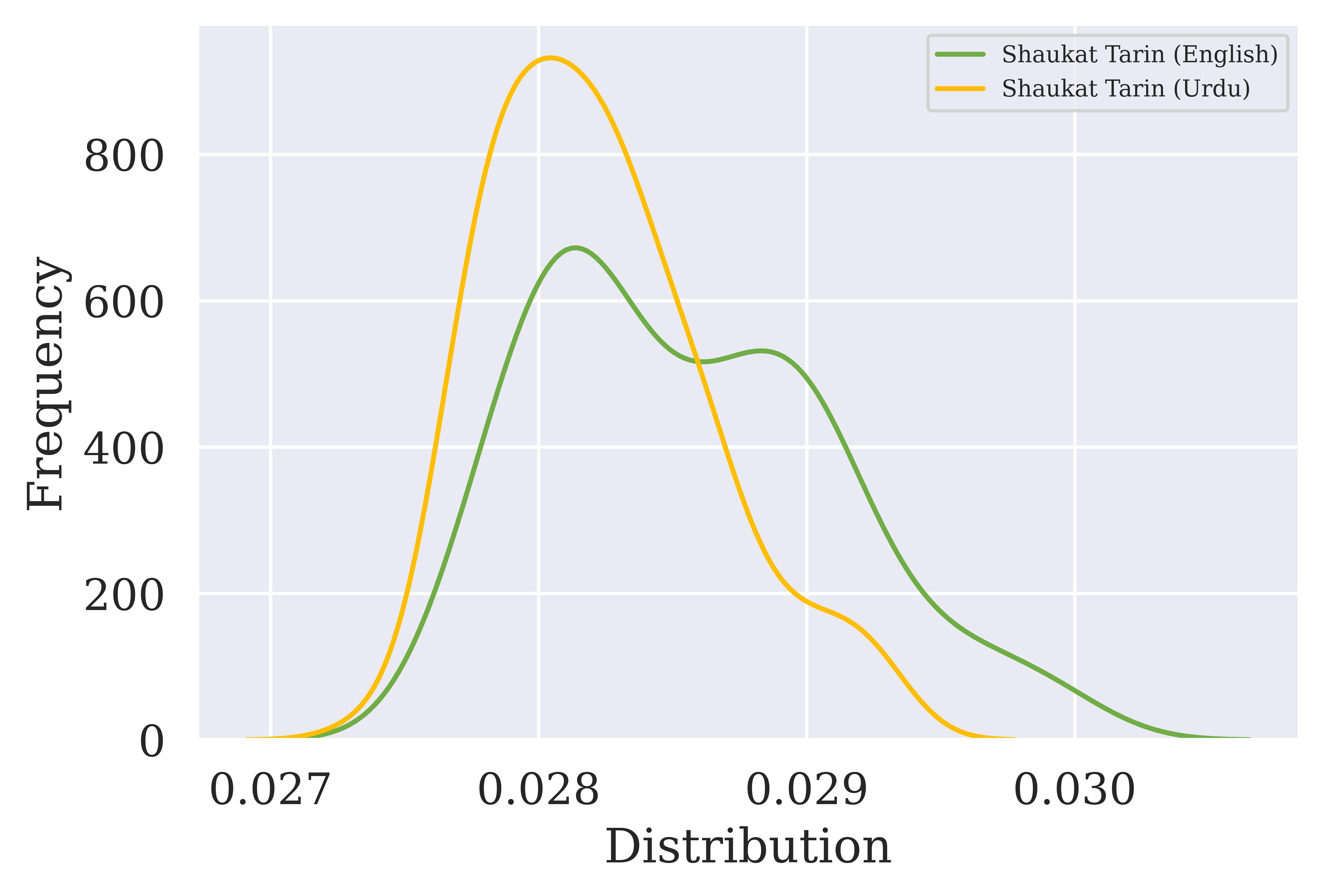}}

\caption{Features distribution of English and Urdu test sets to show domain gap. First row represents feature distribution of English and Urdu test sets with a model trained on English language while the second row represents distribution of English and Urdu test sets with a model trained on Urdu language.}
\label{fig:domain-shift}
\end{figure*}

\noindent \textbf{Impact of multiple languages on face-voice association.} MAV-Celeb dataset provides language annotation of celebrities to analyze the impact of multiple languages on cross-modal verification and matching tasks. For example, a celebrity named `Imran Khan' has audio information in Urdu and English languages. Thus, we can a train a model on one language and tested on an \textit{unheard} one.
Table~\ref{tab:lang} shows the performance of face-voice association on cross-modal verification task across multiple languages using \textit{unseen-unheard} and completely \textit{unheard} configuration. The performance is dropped when our face-voice association model is trained on language `A' (English) and tested on a completely \textit{unheard} `B' (Urdu) language. Similarly, Fig.~\ref{fig:nway-urdu-train} and~\ref{fig:nway-eng-train} show the performance of face-voice association on cross-modal matching task with varying gallery size ($n_c$). 
Interestingly, the performance is better than random verification and matching, which is not trivial considering the challenging configuration \textit{unseen-unheard} and completely \textit{unheard} evaluation protocol. 
In order to find the reason of performance drop, we use feature distributions of three classes in test sets belonging to Urdu and English languages, see Fig.~\ref{fig:domain-shift}. 
We observed that the performance degradation is due to different feature distributions of the two languages. In addition, both languages share common characteristics.

\begin{table}[t]
\caption{Cross-modal verification between face and voice across multiple language on various test configurations of \textit{MAV-Celeb} dataset. The down arrow($\downarrow$) represents percentage decrease in performance. \textbf{(EER: lower is better)}}
\begin{center}
\resizebox{0.48\textwidth}{!}{%
\begin{tabular}{|llcc|}
\hline
       &  & \multicolumn{2}{c|}{\textbf{EU}} \\
\hline\hline
Method & Configuration & Eng. test  & Urdu test   \\
             & & (EER)                          & (EER) \\

\hline
\multirow{2}{*}{MAV-Celeb~\cite{nawaz2021cross}} & Eng. train     & 45.1    & 48.3$\downarrow$\tiny7.1    \\
                                       & Urdu train     & 47.0$\downarrow$\tiny6.3   & 44.3    \\
                                       
\hline
\multirow{2}{*}{Ours} & Eng. train     & 31.0    & 36.6$\downarrow$\tiny17.4    \\
                                       & Urdu train     & 44.4$\downarrow$\tiny41.7   & 31.2    \\

\hline
\end{tabular}}
\end{center}

\label{tab:lang}
\end{table}

\section{Conclusion}
We proposed a light-weight module, namely FOP, for F-V association task. It attempts to harness the best in both face and voice modalities through attention-based fusion and clusters the fused embeddings based on their identity-labels via orthogonality constraints. We integrated this module in a two-stream pipeline, used for extracting face and voice embeddings, and the resulting overall framework is evaluated on a large-scale VoxCeleb$1$ dataset for F-V matching and verification tasks. Our method performs favourably against the existing state-of-the-art methods and proposed FOP outperforms competitors both in accuracy and efficiency. Moreover, we study the performance of our method in the challenging setting of F-V association across multiple languages. Results indicate the opportunity of developing more effective F-V methods that are robust to varying feature distributions in cross-modal verification and matching tasks.

\bibliographystyle{main}
\bibliography{main}

\end{document}